\documentclass[12pt]{article}

\usepackage{scicite}

\usepackage{times}

\usepackage{makecell}

\usepackage{graphicx}
\graphicspath{{figures/}} %Setting the graphicspath

\newenvironment{sciabstract}{%
\begin{quote} \bf}
{\end{quote}}

\newcounter{lastnote}

\title{Optimality Principles in Spacecraft Neural Guidance and Control}

\author
{Dario Izzo,$^{1\ast}$ Emmanuel Blazquez,$^{1}$ Robin Ferede$^{2}$\\
Sebastien Origer$^{2}$, Christophe De Wagter$^{2}$, Guido C.H.E. de Croon$^{2}$
\\
\normalsize{$^{1}$Advanced Concepts Team of the European Space Research \& Technology Centre}\\
\normalsize{Keplerlaan 1, 2200 AG Noordwijk, NL. }\\
\normalsize{$^{2}$Micro Air Vehicle Lab of the Faculty of Aerospace Engineering}\\
\normalsize{Delft University of Technology, 2629 HS Delft, NL.}\\
\\
\normalsize{$^\ast$To whom correspondence should be addressed; E-mail:  dario.izzo@esa.int}
}

\date{}

\begin{document} 

% Double-space the manuscript.

\baselineskip24pt

% Make the title.

\maketitle

% Place your abstract within the special {sciabstract} environment.

\begin{sciabstract}
 Spacecraft and drones aimed at exploring our solar system are designed to operate in conditions where the smart use of onboard resources is vital to the success or failure of the mission. Sensorimotor actions are thus often derived from high-level, quantifiable, optimality principles assigned to each task, utilizing consolidated tools in optimal control theory. The planned actions are derived on the ground and transferred onboard where controllers have the task of tracking the uploaded guidance profile. Here we argue that end-to-end neural guidance and control architectures (here called G\&CNets) allow transferring onboard the burden of acting upon these optimality principles. In this way, the sensor information is transformed in real time into optimal plans thus increasing the mission autonomy and robustness. We discuss the main results obtained in training such neural architectures in simulation for interplanetary transfers, landings and close proximity operations, highlighting the successful learning of optimality principles by the neural model. We then suggest drone racing as an ideal gym environment to test these architectures on real robotic platforms, thus increasing confidence in their utilization on future space exploration missions. Drone racing shares with spacecraft missions both limited onboard computational capabilities and similar control structures induced from the optimality principle sought, but it also entails different levels of uncertainties and unmodelled effects. Furthermore, the success of G\&CNets on extremely resource-restricted drones illustrates their potential to bring real-time optimal control within reach of a wider variety of robotic systems, both in space and on Earth.
\end{sciabstract}
\section*{Introduction}

\begin{figure*}[ht]
\centering
\includegraphics[width=13cm]{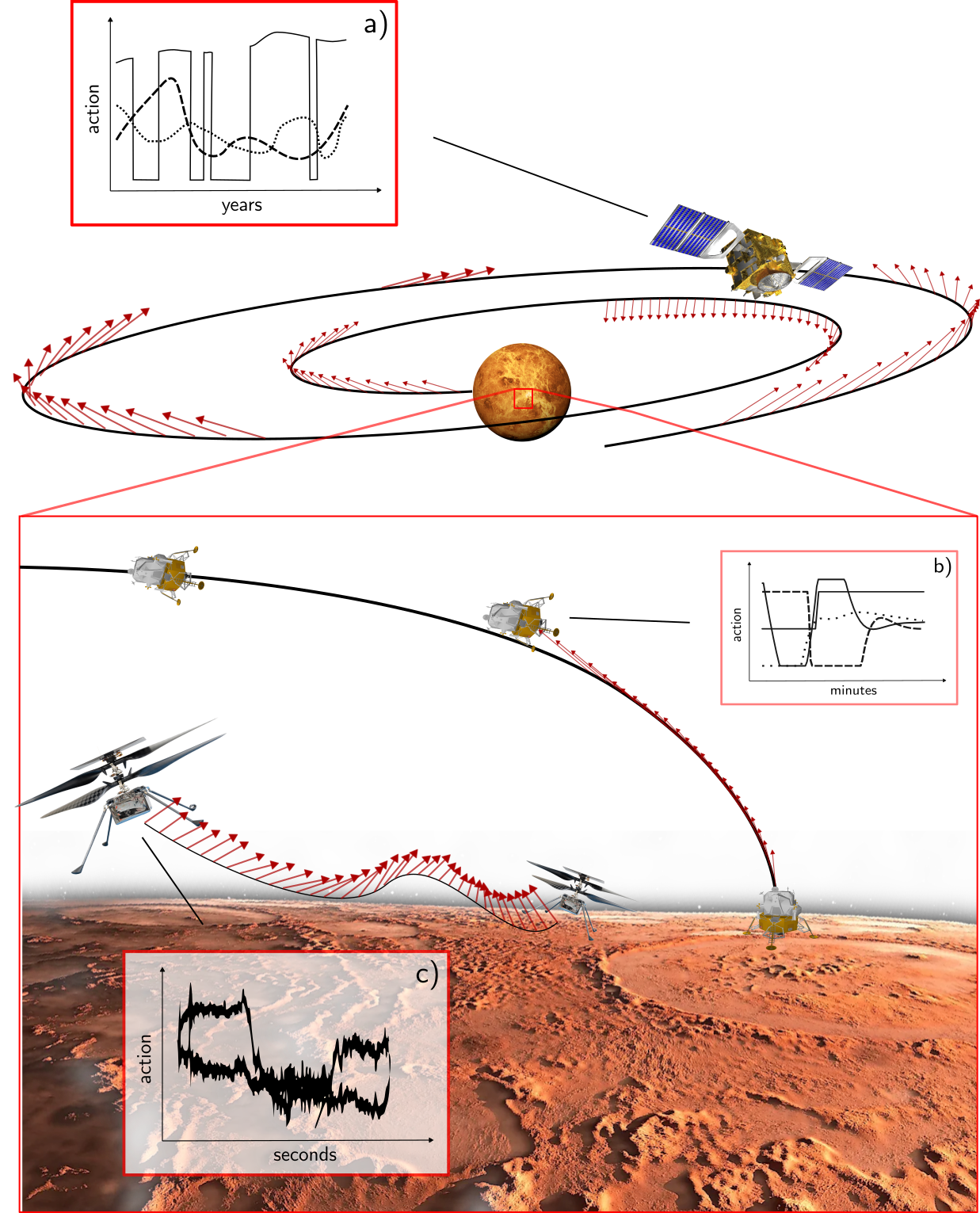}
    \caption{\textbf{Optimality principles determine the decision-making during different phases of exploration missions.} \textbf{a)} During an interplanetary phase, the spacecraft dynamics is well identified. Uncertainties are limited and the departure from a theoretical mass optimal guidance is of less importance also thanks to the relatively slow dynamics involved. \textbf{b)} During a landing phase, according to the specific mission profile, the adaptiveness and robustness of the planned actions have a larger impact on the mission success, also considering that human operators are typically too far away to allow re-planning within an acceptable timeframe. \textbf{c)} During a planetary exploration phase (e.g. rovers or flying drones) uncertainties are larger and optimality principles such as careful use of available onboard energy need to be embedded into highly disturbed and fast dynamics.}
    \label{fig:1}
\end{figure*}

The design of a space exploration mission heavily relies on optimality principles. Given the absence of a safe harbor in space, every onboard resource, including propellant mass, available energy, and computing capabilities, must be utilized parsimoniously to ensure the highest possible mission return. Even with safety margins built into the mission plan, executing sub-optimal plans can result in the failure of the entire mission. To address this challenge, optimal control models have been developed that capture the optimality principles relevant to different mission phases and translate them into elaborate system behaviours. In practice, the optimal guidance profile that follows the application of abstract optimality principles is carefully derived on the ground, well ahead of the mission launch, for most flown and planned missions.
The plan is then uploaded onboard and acted upon by the dedicated control system that, using the specific actuators available, tracks the planned profile continuously canceling incurred deviations. 
This approach is common to interplanetary trajectory phases, landing phases, surface exploration phases, formation flying missions and more (see Figure \ref{fig:1}), thanks to the abstract nature of the well-established optimal control theory which allows to capture different dynamics, actuator models and timescales well. 
It also has the advantage of being a well-tested and validated approach with a history of  successfully embedding optimality principles in the on-board control system allowing missions to meet their requirements. 

This common approach is, however, known to be sub-optimal. It violates a
‘minimal intervention’ principle, well described by \emph{Todorov} and \emph{Jordan} \cite{todorov2002optimal}: efforts in correcting deviations from an average path should be only made when interfering with the task performance. In our context, in case of a deviation from the pre-planned trajectory, the onboard controller should not try to steer back the system to it as the new situation may require other actions for optimality. The approach also carries higher risks in missions where significant unmodeled effects, noise, uncertainties, and unforeseen events are present, which could result in strong deviations from the original plan. In such cases, a new optimal guidance profile may need to be developed on the ground and uploaded to the spacecraft, greatly hindering its autonomy. 

This raises the question of why space missions do not continuously re-plan and compute the optimal guidance profile onboard, for example, by employing Model Predictive Control (MPC). Recent efforts have been made to develop MPC approaches for various mission profiles, aiming to overcome these limitations.
MPC has been studied in the last decades as a promising control approach for aerospace systems (see \emph{Eren et al.} \cite{eren2017model} for a review of MPC in this context) allowing to automate on-board the task of transforming high-level optimality principles into actions. 
It relies on the availability of numerical methods to solve reliably some form of an optimal control problem, starting from the information on the current system state and the time. 
This online optimization returns an optimal sequence of open-loop predicted actions, the first of which is considered the best current control action. 
Despite great advances in the associated numerical techniques and theory\cite{accikmecse2013lossless}, the uptake of modern MPC approaches in space missions remains limited by the available on-board computational capabilities and the reliability of the existing numerical solvers. 

\begin{figure*}[ht]
\centering
\includegraphics[width=\textwidth]{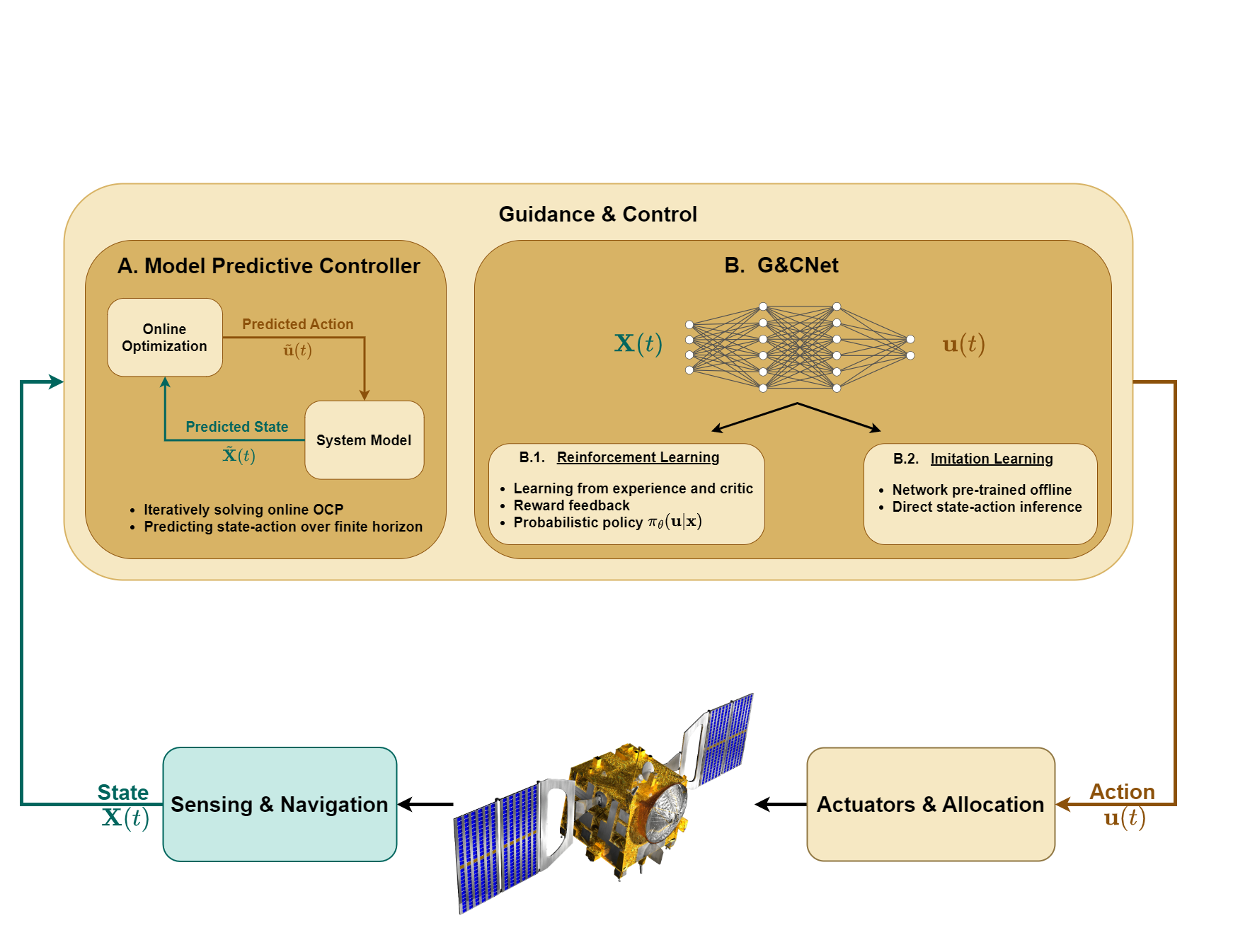}
    \caption{\textbf{G\&CNets have a similar role to Model Predictive Control in the architecture of an autonomous mission.} \textbf{A)} MPC iteratively solves on-board optimal control problems predicting state and actions over a defined time horizon, based on an existing system model. 
    This results in possible optimality guarantees with full predictive information, at the expense of significant on-board computational burden determined by the complexity of the system model and optimal control problem to be solved. 
    \textbf{B)} A G\&CNet inference directly transforms the system state into actions. 
    \textbf{B.1)} When trained using RL, an agent learns from experience the final probabilistic policy, based on a critical reward-feedback loop with the environment. 
    The resulting architecture can be resilient to stochastic disturbances but is often based on engineered reward functions that depart from the original optimality principle assigned. 
    \textbf{B.2)} When trained via supervised learning the network captures a clear optimality principle in its structure, directly inferring optimal actions from the state feedback at high frequency. Such a solution allows for fast direct inference with limited hardware requirements and is possibly subject to instability and lack of robustness when the state falls outside the set used to train the network.
    \label{fig:2}}
\end{figure*}

A different, albeit related, approach appeared more recently in the robotics as well as in the aerospace field.
It is loosely inspired by models that interpret sensorimotor action in humans in terms of optimal control theory \cite{todorov2004optimality}, thus suggesting how optimality principles, in the mathematical sense, are deeply embedded directly in the neuro-musculo-skeletal system.
The new approach attempts to mimic this structure in the guidance and control architecture of space systems and thus to train a deep artificial neural network (DNN) to represent directly the relation between the system state and its action under some predefined task and optimality principle. 
In this context, the network depth refers to the use of multilayer perceptrons that are capable, already with a few layers, of a high representativity \cite{sanchez2018real, tailor2019learning}.
Following this approach, the guidance and control blocks of a typical space mission control architecture are substituted with one, end-to-end, DNN which does not need any form of a reference trajectory and is thus also called a Guidance and Control Net or G\&CNet \cite{izzo2019survey, zavoli2020reinforcement}.
In comparison to the MPC approach, the step of having to solve on-board an instance of the optimal control problem is bypassed and substituted by a single, computationally less expensive, neural inference (see Figure \ref{fig:2}).
Most of the desirable properties of MPC \cite{eren2017model} are instead retained, while the new architecture requires orders of magnitude fewer on-board computational resources. 
Ongoing research in this emerging approach is focused on designing an efficient procedure to train the weights of artificial neural networks to accurately represent the desired task execution. 

In this review, we discuss the use of G\&CNets as a viable and promising path towards achieving on-board optimal decision-making in different space mission scenarios. 
We then discuss two main avenues available for training such networks in the context of space missions: (a) imitation learning, also known as behavioral cloning (essentially supervised learning) and (b) reinforcement learning. 
After evaluating the current state-of-the-art for both approaches, we conclude by proposing drone flight racing as a safe stepping stone towards the on-board implementation of such networks for real missions. 
We finally present concrete examples of the capabilities of embedded G\&CNet implementations using drones as a model system.

\section*{Optimality in guidance and control of space systems}

Space systems are typically well characterized and tested extensively on the ground prior to launch. Their mass, inertia tensor, flexibility, and thrusters are all subject to thorough testing, and precise models of the entire system behavior are developed well in advance of the mission launch. 
However, in some cases, the development of the control system of a space system still requires the consideration of stochastic terms in its dynamics. 
Uncertainties can arise from unmodelable (or difficult to model) effects originating either from the system itself or from the environment in which it is designed to operate. 
Examples of the former include fuel sloshing \cite{Vreeburg2005, Reyhanoglu2011}, mis-thrust events \cite{Servidia2005, Cai2008, Zhou2020}, or corrupted sensor measurements \cite{Yin2016}. Uncertainties coming from the environment arise, for example, in deep space missions to asteroids where the body shape is known only to a limited extent during most mission phases \cite{Melman2013, Ren2015}, or in situations where the atmosphere of some celestial body plays a role such as re-entry, aero-capture or exploratory drone flights \cite{Grip2018, Grip2019, Balaram2021}.
Uncertainties about solar activity or other space environment quantities  also result in unmodeled effects. 

\begin{figure*}[ht]
\centering
\includegraphics[width=\textwidth]{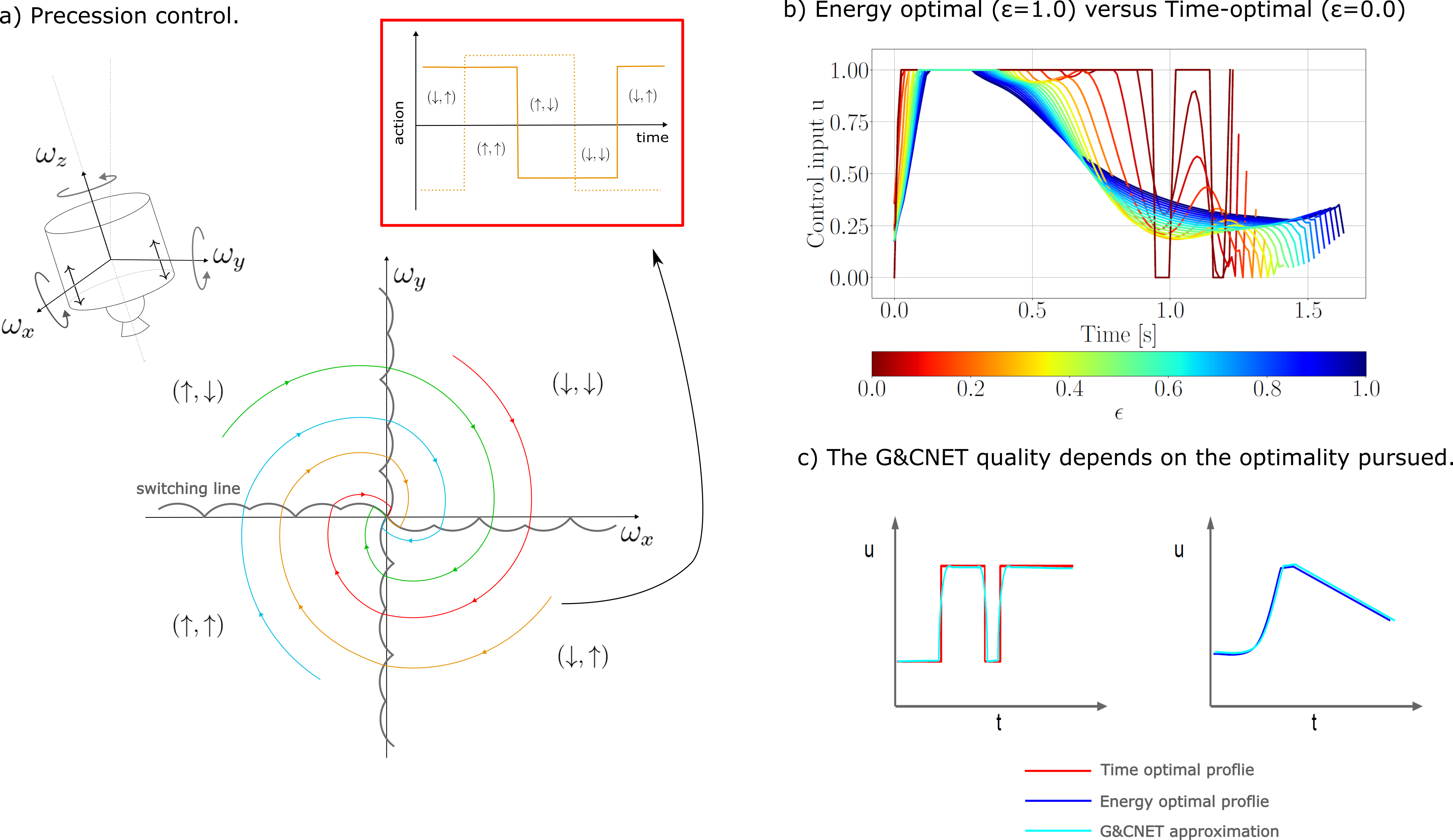}
    \caption{\textbf{Challenges in approximating optimal feedback with a G\&CNet.} \textbf{a)} Optimal control tasks can have very high-dimensional solutions. Already in simple precession control, a complex structure emerges. In this case, the task is to lead in the shortest possible time, a precessing satellite to a uniform rotation around its symmetry axis, thus canceling the components $\omega_x, \omega_y$ of its angular velocity. The resulting deterministic optimal control problem is one rare case where an analytical solution can be derived, allowing us to peek into the structure of the optimal policy over the entire state space. According to the values of $\omega_x, \omega_y$ the thrusters are switching direction in correspondence of a complex and discontinuous switching line. Resulting time-optimal trajectories are shown in color.\textbf{b)} The optimality principle pursued affects the resulting control profile and its gradient. Here the optimal control commands from energy-optimal to time-optimal quadcopter landings are shown \cite{origer2023guidance}. \textbf{c)} Smooth control profiles result in smaller errors compared to non-smooth bang-bang profiles when approximated by a G\&CNET.
    \label{fig:3}}
\end{figure*}

In any case, ignoring the stochastic contribution and using the sophisticated mathematical framework for deterministic optimal control emerged from the work of \emph{Pontryagin} \cite{pontryagin} offers a starting point in understanding the underlying structure rising from chosen optimality principles. 
Under this framework, it is well known how the optimal feedback is a discontinuous and non-linear function of the system state already for most simple low-dimensional and linear systems \cite{pontryagin}. An informing example is that of the time optimal precession angle control of a spinning satellite \cite{izzo2004internal} where the analytical solution is available and allows exceptionally, to observe the optimal control switching structure in the full state space, revealing the extremely nonlinear and discontinuous nature of such a function (see Figure \ref{fig:3}).
This is the case for time optimality, but also and to a larger extent for higher dimensional cases and propellant mass optimality: the primary optimality principle for most deep-space mission phases. 
In deep-space missions, thrust is mostly achieved through the ejection of propellant mass which causes the spacecraft to lose mass and accelerate in the opposite direction. 
This specific form of control for mass varying systems creates a distinct structure of the resulting optimal actions which is the subject of a large body of works from the aerospace community (see for example \cite{conway2010spacecraft}). 
Thrust is also often modeled as a sequence of impulsive velocity changes rather than as a continuous action, and dedicated concepts such as the primer vector introduced by \emph{Lawden} \cite{lawden} have been used to expand upon Pontryagin work and help to answer complex questions on the resulting intricate structure of the optimal sequence of impulsive maneuvers \cite{edelbaum1967many, prussing1995optimal, carter2000necessary, taheri2020many}. 

When stochastic terms are considered in the system dynamics, the mathematical structure of the optimal control problem undergoes a significant change. As a result, Pontryagin theory is no longer applicable, and Bellman's dynamic programming methods \cite{bellman1965dynamic} offer a more appropriate tool based on the concept of the optimal cost-to-go or value function \cite{todorov2006optimal}. This added complexity is captured by a set of nonlinear, second-order, partial differential equations known as the Hamilton-Jacobi-Bellman equations \cite{bardi1997optimal}. Finding a solution to these equations yields the value function and thus the optimal policy, but even in the simplest cases, it can be challenging, particularly for problems of interest in this context.

Most importantly, Belmann's mathematical framework allows for demonstrating the existence and uniqueness of solutions in terms of the value function, and thus the existence (outside of singular corner cases) of optimal feedback in the form $\mathbf u^*(\mathbf x)$ where $\mathbf u^*$ denotes the optimal feedback and $\mathbf x$ denotes the system state. In summary, for any (deterministic or stochastic) task, the current system state and an associated optimality principle are sufficient to decide upon the action to be taken. However, such an optimal state-action mapping is nonlinear, discontinuous, and has an extremely intricate differential structure.

\section*{Embedding optimality principles into neural models}
Artificial neural networks are highly versatile in their ability to approximate complex and discontinuous functions, even in their simplest shallow feed-forward architecture, as recently re-discussed in depth by \emph{Calin} \cite{calin2020universal}. 
They have been demonstrated to represent the optical characteristics of complex 3D scenes with remarkable detail, as evidenced by the work of Mildenhall et al. \cite{mildenhall2020nerf}, as well as to model the gravitational field of irregular bodies in the solar system with a precision exceeding that of classical methods, such as spherical harmonics expansions \cite{izzo2022geodesy}.
Therefore, DNNs are an obvious choice for representing the complex structure of the optimal policies $\mathbf u^*$ (or the value function $v$) often needed in space missions.
A DNN (here denoted generically with $\mathcal N_{\theta}$) can formally approximate, within some tolerance $\epsilon$, a parametric optimal feedback of the form $\mathbf u^*(\mathbf x, \mathbf p) = \mathcal N_{\theta}(\mathbf x, \mathbf p) + \epsilon$, where the parameters $\mathbf p$ may capture different tasks, objectives and environmental properties, such as constraints and unknown gravitational effects.
As a result, the biases and weights $\theta$ of the neural model incorporate multiple optimality principles, which are subsequently converted into control commands through onboard inference. This inference process is becoming increasingly efficient due to the development of dedicated AI accelerators for on-the-edge space systems, leading to the creation of new dedicated hardware\cite{furano2020towards}. AI-focused processors were embarked in the ${\Phi}$-Sat-1 \cite{PhiSat1} mission, field-programmable gate arrays (FPGAs) on OPS-SAT \cite{OPSSATOnboardAI} and HYPSO-1 \cite{danielsen2021self, pitonak2022cloudsatnet}, and graphics processing units (GPUs) are being considered for future non-mission-critical applications \cite{OrbitalEdge, UnibapSpaceCloud}. 
The use of DNNs for onboard systems with limited computational resources, such as spacecraft, cubesats, and planetary drones, has been only recently proposed in the context of space missions and is attracting the attention of a growing number of scientists.
The term G\&CNets (guidance and control network)\cite{izzo2019survey, zavoli2020reinforcement}, introduced in early studies at European Space Agency, is here used to indicate such DNNs promising to replace traditional control and guidance architectures in future space missions. 
G\&CNets main attractiveness stems from their promise to bypass problems connected to the onboard solution of optimal control problems, typical for example of classic MPC schemes \cite{eren2017model}, at the cost of introducing the need of pre-training robust neural models on the ground. 
Trivially, in the limit case in which both an MPC and a G\&CNets are able to compute the actual true optimal-feedback $\mathbf u^*$ without errors and with similar computational complexity, they are equivalent. 
Two main approaches are mostly being studied in the context of G\&CNets training: supervised learning and reinforcement learning.

\subsection*{Approaches based on Supervised Learning}
The model parameters $\theta$ of the DNN approximating the parametric optimal feedback  $\mathcal N_\theta(\mathbf x, \mathbf p)$ can be learned via a supervised learning approach provided some dataset $\mathcal D :=\left\{(\mathbf x, \mathbf p), \mathbf u^*\right\}$ is available containing optimal state-action pairs for one or more tasks represented by $\mathbf p$. This approach can also be referred to as imitation learning \cite{tailor2019learning} or behavioral cloning \cite{mulekar2023metric} to highlight the control aspect of the dataset to be learned. 
After the successful demonstration over a range of simulated landing tasks \cite{sanchez2018real}, a large number of works independently tested the capabilities of imitating the optimal control in diverse space contexts such as lunar and Mars landing \cite{mulekar2023metric, furfaro2018deep}, irregular asteroid landings \cite{cheng2020real}, low-thrust missions and orbital transfers \cite{izzo2021real, IZZO2022, li2019neural}, solar sailing \cite{cheng2018real}, proximity operations \cite{federici2021deep}, drone flights \cite{tailor2019learning} and mis-thrust problems \cite{rubinsztejn2020neural}. While achieving convincing results in many of the cases reported, much research still needs to be done to tackle the issues connected to the imitation learning approach: the efficient creation of a dataset $\mathcal D$, the verification of requirements on the resulting onboard control system \cite{izzo2020stability}, and the inclusion of mechanisms able to cope with unmodeled components\cite{rubinsztejn2020neural}.

\begin{figure}[ht]
  \centering
    \includegraphics[width=13cm]{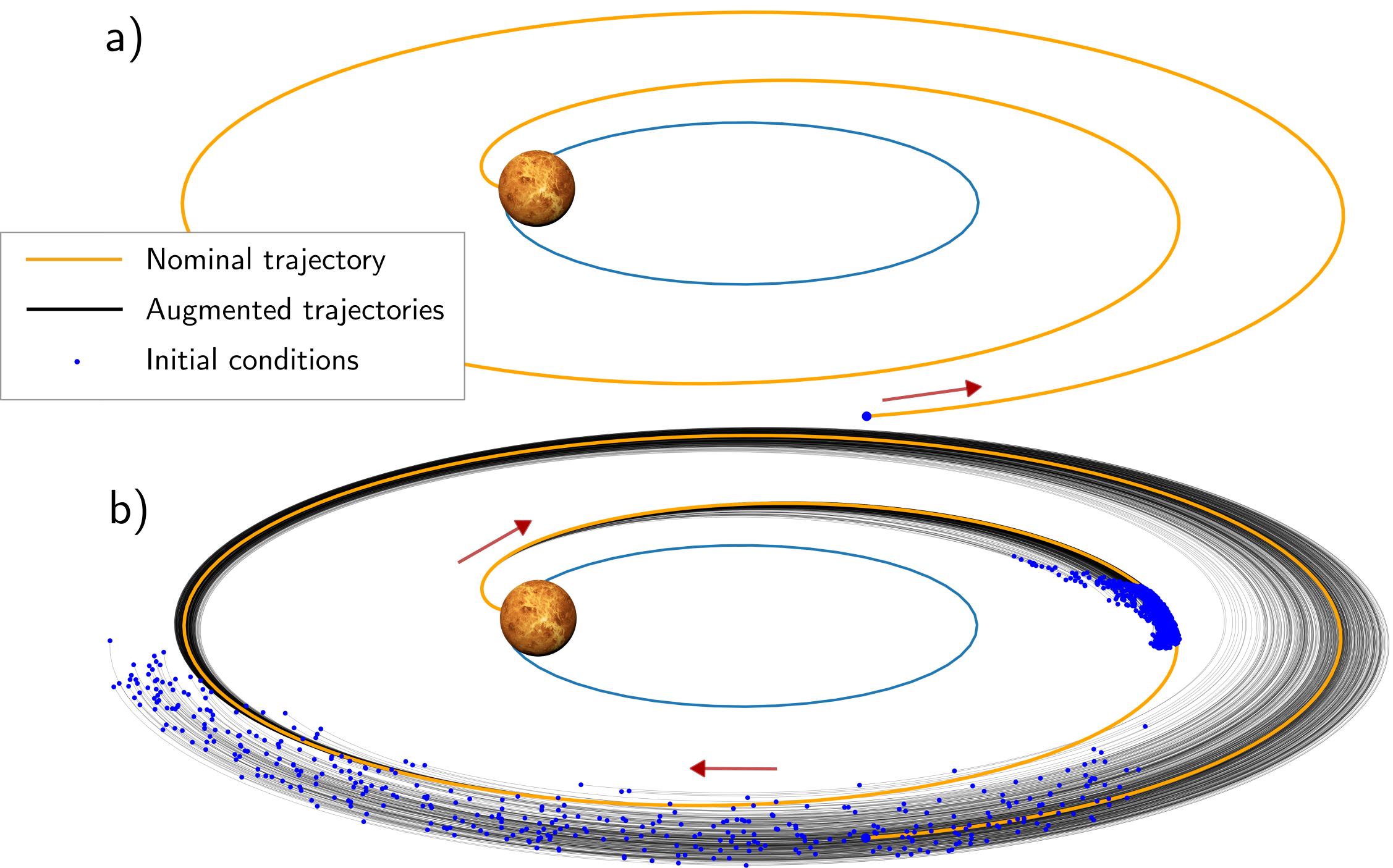}
  \caption{\textbf{The backward generation of optimal examples (BGOE) technique, allows to generate orders of magnitude larger datasets by perturbing one nominal solution.}  \textbf{a)} The nominal solution for the case of a time-optimal transfer from the asteroid belt to Earth (visualized in a rotating frame). \textbf{b)} Two bundles of 200,000 optimal trajectories found applying BGOE to the nominal solution. 
  Larger perturbations of the nominal trajectory result in better coverage of conditions close to the Earth (short bundle), which reduces the likelihood that the spacecraft lands outside of the training data \cite{IZZO2022}.
  For comparison, the generation of all 400,000 trajectories uses the same numerical resources used to generate one nominal optimal solution. }
  \label{fig:4}
\end{figure}

The creation of a dataset $\mathcal D$ requires running numerical optimal control solvers over a large set of initial states. This demanding computational effort can be alleviated by the use of techniques leveraging the proximity to previous solutions found (such as homotopy or continuation \cite{trelat2012optimal}), but remains a limiting factor, although not one burdening the on-board inference speed.
As a consequence, the number of optimal trajectories needed to build $\mathcal D$ and test the capabilities of an imitation learning approach to G\&CNets, have been, so far, mostly limited to the order of tens of thousands \cite{cheng2018real, li2019neural, rubinsztejn2020neural, mulekar2023six}.
Small datasets are unable to harness the full potentials of a deep supervised learning pipeline and only allow for partial investigations of the potentials of G\&CNets, restricting possible results to only small portions of the state space.
In cases where Pontryagin's theory can be used to derive necessary conditions for optimality via the introduction of a two-point boundary value problem defined on the augmented dynamics (indirect methods) a technique called Backward Generation of Optimal Examples - or BGOE -  (see Figure \ref{fig:4}) has been recently introduced to allow the creations of  datasets $\mathcal D$ orders of magnitude larger than what is otherwise possible. BGOE has been used in the context of studies on Earth-Venus mass optimal interplanetary transfers \cite{izzo2021real}, as well as time-optimal asteroid belt mining missions \cite{IZZO2022}, showing promising results and allowing the creation of public datasets containing millions of optimal trajectories rather than only a few thousand.

Studies on the use of BGOE are still preliminar, and the technique cannot be used in general, for example in contexts where indirect optimization methods fail. 
As a consequence, most of the past works on imitation learning for G\&CNets suffer from an insufficient representativity of the trained network, resulting in the system state propagating outside the training set and thus the onboard closed-loop becoming unstable.
In the presence of strong perturbations and uncertainties, this can be the case even if the dataset size is appropriate.
To tackle this issue, approaches of data augmentation implementing different variations of the DAGGER \cite{ross2011reduction} technique have been proposed and used \cite{furfaro2018deep, mulekar2023metric, mulekar2023six, rubinsztejn2020neural}, but also criticized as they increase the burden of generating additional optimal state-action pairs.

Finally, the question of how many different tasks and optimality principles can be embedded in a single DNN via imitation learning remains open as work addressing the issue has only been carried out preliminarily for interplanetary trajectories \cite{sprague2020learning} and drones \cite{origer2023guidance, Ferede2023Adaptive}. 
These early results show how the introduction of multiple tasks and environment variables encoded in the extra parameters $\mathbf p$ seem to help in regularizing the discontinuous behavior resulting from aggressive optimality principles as well as in coping with unforeseen unmodeled external perturbations (see the later section on onboard drone flight implementation). 

\subsection*{Approaches based on Deep Reinforcement Learning}
A second approach suitable to learn the parameters $\theta$ of a G\&CNet is deep reinforcement learning (DRL). DRL captures a large variety of approaches and numerical methods concerned with the problem of an agent learning from experience, and not from expert demonstrations, a policy to perform a specified task in its environment. The policy -- often indicated with the symbol $\pi_\theta(\mathbf u | \mathbf x)$ -- is represented by a DNN and returns the probability to choose the control $\mathbf u$ given the agent state $\mathbf x$. 
In a deterministic setting such a policy, when optimal, must correspond to the solution of the related optimal control problem so that one can formally write $\pi^*_\theta(\mathbf u | \mathbf x) = \delta(\mathbf u - \mathbf u^*(\mathbf x))$ where the Dirac delta has been employed to indicate certainty over choosing the optimal feedback $\mathbf u^*$. In a stochastic setting, the DNN typically outputs some statistical property of the policy, most commonly its mean value -- the variance being often considered constant. 
In order to use DRL to train the parameters of a G\&CNET, the continuous control problem has to be modeled as an agent that learns through a sequential decision-making process, a Markov Decision Problem (MDP), where the simulated environment includes all types of uncertainties relevant to the particular mission phase considered.
Early works \cite{gaudet2012robust, willis2016reinforcement}, for example, applied this approach to the problem of controlling a spacecraft hovering over an asteroid with an uncertain gravity field and considering a stochastic dynamics perturbed by solar radiation pressure acceleration as well as accounting for sensor noise. 
The approach was later extended and refined to interplanetary transfers \cite{miller2019interplanetary, zavoli2020reinforcement}, rendezvous and docking \cite{yuan2022deep, oestreich2021autonomous}, planetary landing problems \cite{furfaro2017waypoint}  as well as drone flights \cite{song2021autonomous, azar2021drone}. 
% SHall we here mention PPO and DPS as the mostly used methods?
In all these cases the trained DNN is suitable for on-board utilization and proved to be robust to different levels and types of stochastic effects. While very promising, the DRL approach has much to prove in terms of actual optimality. 
Optimality principles as well as terminal and path constraints are all encapsulated in the so-called reward function driving most of the agent learning. 
Engineering an appropriate reward function turns out to be problematic in most cases and, when successful, it is unclear to what optimality principle it corresponds to. 
This is not necessarily an issue when optimality with respect to an assigned objective is not of primary concern, but it is nevertheless something to be well aware of in most scenarios of interest to space missions.
Not surprisingly, the sub-optimality resulting from the DRL approach is noted and quantified by the authors of successful implementations \cite{zavoli2020reinforcement, yuan2022deep, miller2019interplanetary}. 
In this context, it may be argued how modern DRL efforts do not seek to optimize a given objective function better than methods based on optimal control theory, but that instead they intrinsically define a better objective. We agree that this may be indeed the case, but only for systems where uncertainties are prevailing and approaches grounded on the underlying optimal control theory (i.e. Bellman equations) fail to provide sufficient performances. 
A common criticism toward the DRL approach derives from its intense computational requirements that can result in three orders 
of magnitude larger computational times to find a policy with respect to, for example, running one single optimal control solver \cite{zavoli2020reinforcement}. Finally, to the best of our knowledge, there are no attempts to include in the DRL framework the possibility to learn a class of policies and thus embed in the same deep network multiple space tasks, objectives and constraints. 

\section*{Testing on Flying drones}
The field of onboard guidance and control of space systems using deep neural networks (DNNs), specifically the G\&CNet architecture, is in its infancy. To gain acceptance within the aerospace community as a potential improvement over current guidance and control (G\&C) schemes, convincing evidence of its embedded capabilities is necessary, as well as of the possibility to offer guarantees on the resulting system behavior. With much of the current research on G\&CNets focused on simulations, concerns are raised that the reality gap may be overlooked. 

Drone racing platforms share similarities with space systems, as they too have limited onboard resources and require careful optimization of their utilization. The time optimality principle in drone racing, coupled with the complex dynamics of the drones, provides a valuable comparison to the mass optimality principles preeminent in space missions.
An ongoing effort is underway to integrate G\&CNets into quadcopter flight control systems, aiming to achieve end-to-end control with optimal flight time and to increase the trust in the use on-board spacecraft.
Building upon research that employed simplified quadcopter models in simulations \cite{sanchez2018real, tailor2019learning} and performed real-world flight tests where G\&CNets were utilized for longitudinal control \cite{li2020aggressive}, tests have been conducted where a high-dimensional quadcopter model with sixteen degrees of freedom was employed to compute the optimal-feedback for the networks to imitate.
The networks were then trained for end-to-end control, which involved sending motor commands directly without any intermediate controllers \cite{Ferede2023Adaptive, origer2023guidance}.
This approach granted the network complete control authority, allowing it to handle actuator saturation directly and without being constrained by an inner-loop controller as in previous attempts. However, this also posed new challenges, as the time optimality principle pursued resulted in aggressive maneuvers that were highly dependent on the network's ability to accurately replicate the ground truth optimal actions. Due to the short dynamical timescale of the quadcopter flights considered (on the order of seconds), even minor discrepancies in the network's performance significantly impacted the overall control accuracy.
In hover-to-hover flights, unmodeled moments caused the drone to take a sub-optimal path, diving down and overshooting its target. Additionally, modeling errors accumulated and caused deviations from the optimal plan, which destabilized the trajectory \cite{Ferede2023Adaptive}.

\begin{figure}[ht]
  \centering
    \includegraphics[width=13cm]{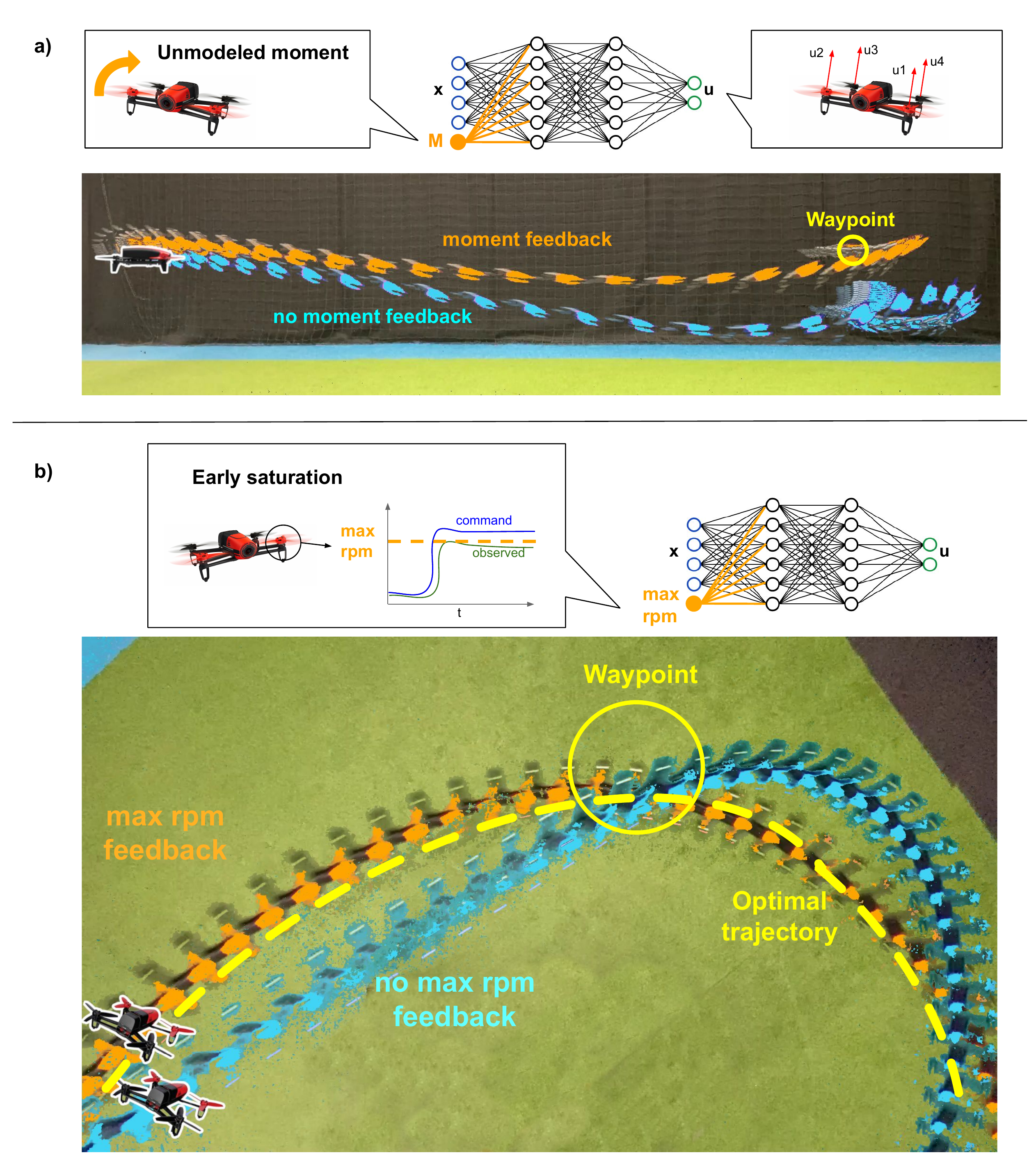}
     \caption{\textbf{G\&CNets robustness to model mismatch.} Two examples of real autonomous flights of a Parrot\textsuperscript{\texttrademark} AR drone 2.0 in the TU Delft Cyberzoo: \textbf{a)} Unmodeled moments are detected in real-time on-board and fed back to the G\&CNet. \textbf{b)} Unexpected early saturation of the motor revolutions per minute is present. The saturation is estimated on-board and fed back to the G\&CNet. In both cases, the improvements with respect to the non-parametric version are evident. The initial drone position is also shown with a white border.}
  \label{fig:5}
\end{figure}

Despite the difficulty in modeling the system dynamics, sensor measurements onboard can be used to detect discrepancies between predicted and actual forces and moments.
This suggests to modify the G\&CNet adding to its inputs estimated unmodeled effects. A modified network $\mathcal N_\theta(\mathbf x, \mathbf p)$ can then trained to approximate the solution of a parametric optimal control problem where $\mathbf p$ represents the discrepancy between the modeled dynamics and the observed behavior. By comparing the predicted and measured moments onboard and using their difference as input to the network\cite{Ferede2023Adaptive}, G\&CNets have been shown to adapt to unexpected moments delivering a significant improvement in stability and optimality, as illustrated during real flights of a Parrot\textsuperscript{\texttrademark} AR 2.0 drone in Figure \ref{fig:5}a.

In \cite{origer2023guidance}, a similar approach was adopted, where the G\&CNet was modified to instead handle unexpected actuator saturations. It was demonstrated that the knowledge of the actuator's limits was crucial to remain on the optimal path during high-speed successive waypoint flights. By estimating a model parameter onboard (in this case, the maximum motor revolutions per minute) and feeding it back into an additional neural network input, G\&CNets can thus cope with unmodeled effects, as shown in Figure \ref{fig:5}b.

\begin{figure}[ht]
  \centering
    \includegraphics[width=\textwidth]{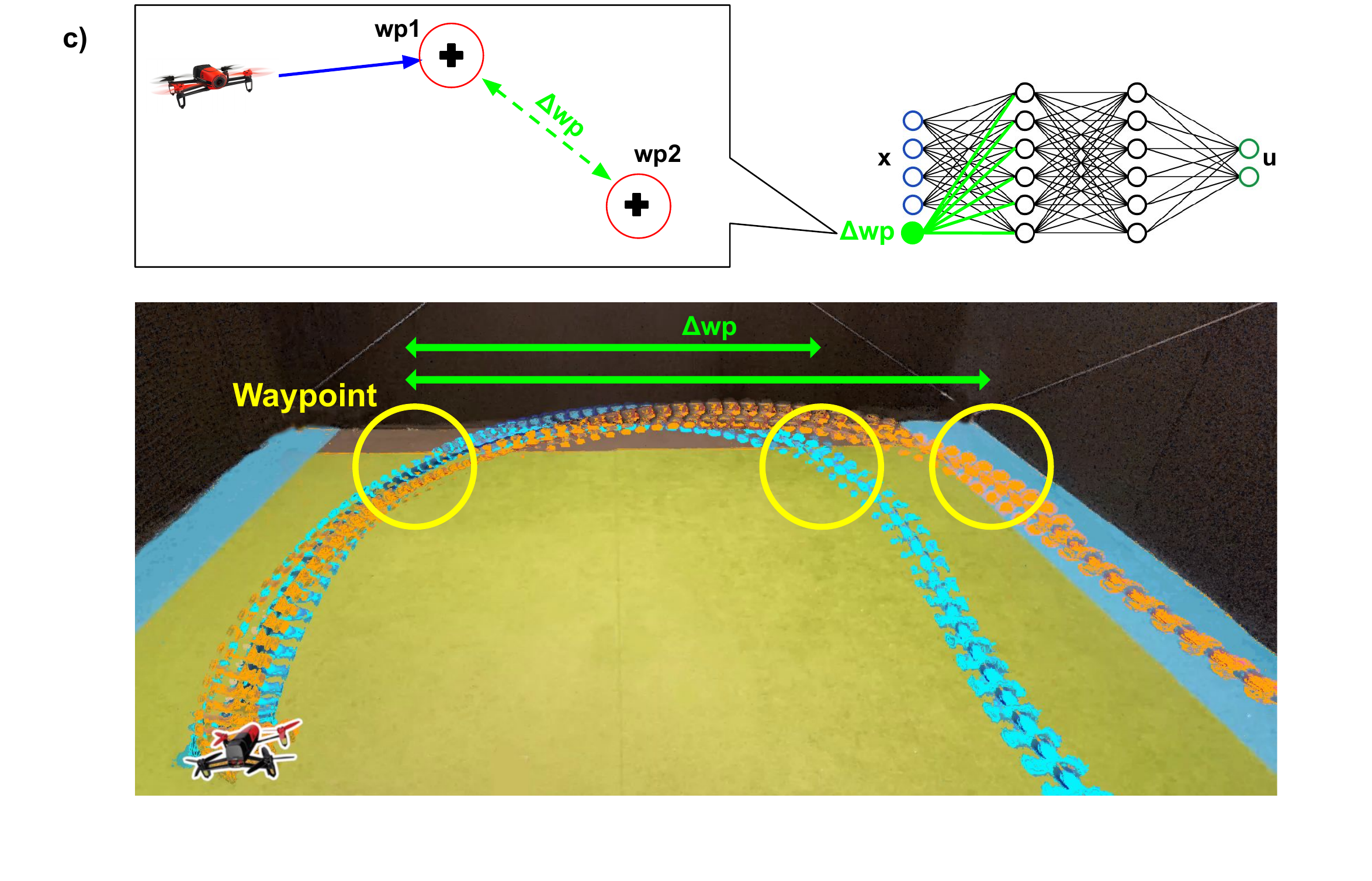}
  \caption{\textbf{Embedding multiple task in one G\&CNet.} Real  autonomous flights of a Parrot\textsuperscript{\texttrademark} AR drone 2.0 in the TU Delft Cyberzoo. During this specific test, the same G\&CNet is shown to have learned to imitate the optimal feedback in two distinct tasks differing in the final waypoint position. The position of the next waypoint is fed to the DNN as an extra parameter. The initial drone position is also shown with a white border.}
  \label{fig:6}
\end{figure}

Additional parameters fed into a G\&CNet can also represent different tasks. For instance, in \cite{origer2023guidance}, a G\&CNet is also trained to fly through two waypoints, where the waypoint positions in space, boundary conditions to the optimal control problem, can vary. 
When computing the state-action pairs to use in the training dataset, different distances between waypoints are used, and the network thus learns to represent the solution of the optimal control problem specific to the parametric geometry it receives as additional input. 
During flight, the network can then optimally plan a path through two waypoints, where the second waypoint position is displaced, as illustrated in Figure \ref{fig:6}.

% (Additional challenges)
G\&CNets face significant challenges when highly aggressive or acrobatic maneuvers are pursued on drones. 
The G\&CNets here showcased during in real flights were trained to imitate energy optimal control \cite{Ferede2023Adaptive} or a mixture of energy and time optimal control \cite{origer2023guidance}. Energy optimality was preferred as it resulted in smoother trajectories, which were simpler to learn and execute. However, moving towards more time-optimal trajectories also creates a challenge as the optimal solution approaches a bang-bang profile, which is increasingly difficult to learn (see Figure \ref{fig:3}c), particularly for the small networks used \cite{origer2023guidance}. Moreover, the aggressive commands will steer the drone toward the edge of its flight envelope where the dynamics are most unpredictable. Therefore the reality gap problem remains the biggest obstacle to this approach in the context of aggressive maneuvering. Although the adaptive networks have shown significant robustness improvement, additional alterations are necessary to guarantee more general robustness for modeling errors, sensor noise and delays.

% (Conclustions)
These findings demonstrate that relatively small G\&CNets, in this case a feed-forward neural network with three hidden layers consisting of 120 neurons each, are capable of representing a broad range of optimal control problems and tasks effectively.
The resulting G\&CNets are able to cancel out, rather than accumulate the errors made in-flight, both in simulation and on the real quadcopter. This is in part possible because they can be made adaptive, but also because they can be inferred at a high rate ($450$Hz) in real-time, onboard of the drone (as measured on the Parrot P7 dual-core Cortex A9 CPU).
%The G\&CNets generated by the approach are capable of canceling out errors made in-flight rather than accumulating them, both in simulation and on the real quadcopter.
%This capability is partly due to the fact that the G\&CNets can be made adaptive and partly because they can be inferred at a high rate (450 Hz) in real-time on-board the drone. 
A Dual-core 800 MHz ARM Cortex-A9 processor is also currently orbiting on-board the European Space Agency's OPS-SAT satellite \cite{evans2014ops}, thus making the results achieved of particular interest here as they map closely to the space systems with similar computational capabilities.

%\section*{Future directions and challenges}
%- Joining RL + Imitation Learning
%- Spiking Nets, event-based navigation (neuromorphic) (for fast systems like %%landings)
%- Hardware - onboard spacecraft accelerators (from ESA)  OPS-SAT .... [roadmap to %implementation]

\section*{Future works}

Small neural networks have demonstrated, in several scenarios relevant to space exploration, their ability to capture optimality principles and perform various guidance and control tasks requiring significant levels of autonomy. As the development of specialized accelerators for on-board inference continues to grow \cite{furano2020towards}, there is increasing interest in exploring which additional capabilities can be integrated into larger networks. However, due to the data-intensive nature of deep architectures, the efficient generation of training examples is imperative for realizing the potential of larger networks. End-to-end training of large models, when performed on-board spacecraft, remains an open question, subject to availability of on-board acceleration and the development of distributed learning systems as a key enabling technology.\cite{mcmahan2017} This could result in significant autonomy benefits for future deep space missions, and recent interest in satellite swarms and constellations may rapidly lead to new developments in this field.

The question of whether the training procedure of G\&CNets should imitate optimal examples or utilize a reinforcement learning paradigm remains open and can today only be answered on a case-per-case basis, based on on the user's desired trade-off between computational efficiency, robustness, quest for optimality and interpretability. While exploring the synergies between these two approaches may lead to innovative, hybrid solutions, it is clear that this topic will continue to fuel scientific discussions in the years to come.

Unresolved challenges remain regarding the complete qualification and validation of artificial neural networks when embedded into mission-critical systems like the guidance, navigation and control subsystem. 
To achieve certification, end-to-end neural guidance and control implementations need to demonstrate robustness to stochastic parametric and dynamic system uncertainties, as well as compatibility with modern fault detection, isolation and recovery (FDIR) routines that require both significant explainability and adaptability. 
It is crucial to ensure that neural systems perform as intended under a range of operating conditions and environments while meeting safety and performance requirements comparable with modern day control solutions. 
This involves rigorous testing and validation of the underlying models and algorithms, as well as ensuring that the system can operate reliably in the presence of uncertainties such as sensor noise, communication delays, and other disturbances commonly faced in space applications. 
Fortunately, recent developments in explainable artificial intelligence (XAI) offer new possibilities for developing more comprehensive deep neural models that can provide greater transparency and understanding of their capabilities and limitations, which could ultimately unlock mission-critical certifications.\cite{BARREDOARRIETA202082}

Neuromorphic technologies have recently emerged as promising enablers for on-board edge computing and learning applications in space. These technologies propose novel software and hardware designs inspired by biological neural systems, with a focus on low-power and energy efficiency, which are highly synergetic with space applications. The widespread availability of neuromorphic Dynamic Vision sensors \cite{brandli2014} and chips \cite{davies2018} has led to a recent surge of publications on the use of event-based cameras and spiking neural networks on-board spacecraft \cite{Chin2019, Sikorski2021, McLeod2023, Azzalini2023}, with an event-based sensor having even recently been sent to the International Space Station \cite{McHarg2022}. Of particular interest for G\&CNets is the neuromorphic promise to enable real-time continuous learning, with the potential to revolutionize neural guidance and control capabilities and robustness.

\section*{Conclusions}

In this study, we have reviewed a nascent trend in neural spacecraft guidance and control that leverages an on-board deep artificial neural network (DNN) to capture the correlation between the spacecraft's state and its optimal actions, inspired by the presence of optimality principles in human sensorimotor actions.
The proposed Guidance and Control Net (G\&CNet) architecture is computationally efficient, suitable for real-time on-board processing and has shown great potential in representing complex optimal state-feedback relations in various scenarios of interest while maintaining many of the favorable attributes of more classical control methods. 
Preliminary attempts to build trust in this approach have been presented that successfully embed G\&CNet on-board drones on hardware compatible with modern space CPUs using time-optimal flights as a proxy for a real space guidance and control system.

Looking ahead, we foresee the adoption of on-board neural guidance and control becoming more widespread in the future, as it enables the necessary autonomy to meet the ambitious goals of future space missions, while parsimoniously utilizing on-board available resources.

%\begin{figure*}[ht]
%\centering
%\includegraphics[width=13cm]{figures/figure2_alt.png}
%    \caption{\textbf{Autonomous spacecraft guidance and control architectures using Model Predictive Control or G\&CNETs.} \textbf{A)} MPC iteratively solves on-board optimal control problems predicting state and actions over a defined time horizon, based on an existing system model. \textbf{B)} Using either a reinforcement learning agent or a network derived from imitation learning. \textbf{B.1)} The RL agent learns from experience in maximizing a given policy, based on a critical reward-feedback loop with the environment. \textbf{B.2)} The imitation network is pre-trained offline and captures optimality principles in its structure, directly inferring optimal actions from state feedback at high frequency.
%    \label{fig:2_alt}}
%\end{figure*}

\bibliography{main}
\bibliographystyle{Science}

\end{document}